\documentclass[a4paper,conference]{IEEEtran}
\usepackage{hyperref}
\usepackage{multirow}

\usepackage{amsmath}

\usepackage{cite}

\ifCLASSINFOpdf
   \usepackage[pdftex]{graphicx}
\else
\fi
\usepackage{subcaption}

\usepackage{xcolor}

\begin{document}
%
\title{QuantFace: Towards Lightweight Face Recognition by Synthetic Data  Low-bit Quantization}

\author{\IEEEauthorblockN{Fadi Boutros}
\author{Fadi Boutros$^{1,2}$, Naser Damer$^{1,2}$, 
 Arjan Kuijper$^{1,2}$\\
$^{1}$Fraunhofer Institute for Computer Graphics Research IGD,
Darmstadt, Germany\\
$^{2}$Department of Computer Science, TU Darmstadt,
Darmstadt, Germany\\
 Email: {fadi.boutros@igd.fraunhofer.de}
}

\IEEEauthorblockA{Fraunhofer Institute for Computer Graphics Research IGD,
Darmstadt, Germany\\
Mathematical and Applied Visual Computing, TU Darmstadt,
Darmstadt, Germany\\
 Email: fadi.boutros@igd.fraunhofer.de} 
\and
\IEEEauthorblockN{Naser Damer}
\IEEEauthorblockA{Fraunhofer Institute for Computer Graphics Research IGD,
Darmstadt, Germany\\
Department of Computer Science, TU Darmstadt,
Darmstadt, Germany\\
 Email: naser.damer.boutros@igd.fraunhofer.de}
\and
\IEEEauthorblockN{Arjan Kuijper}
\IEEEauthorblockA{Fraunhofer Institute for Computer Graphics Research IGD,
Darmstadt, Germany\\
Department of Computer Science, TU Darmstadt,
Darmstadt, Germany\\
 Email: arjan.kuijper@igd.fraunhofer.de}}


%


\maketitle

\begin{abstract}
Deep learning-based face recognition models follow the common trend in deep neural networks by utilizing full-precision floating-point networks with high computational costs. Deploying such networks in use-cases constrained by computational requirements is often infeasible due to the large memory required by the full-precision model. 
Previous compact face recognition approaches proposed to design special compact architectures and train them from scratch using real training data, which may not be available in a real-world scenario due to privacy concerns.
We present in this work the QuantFace solution based on low-bit precision format model quantization. 
QuantFace reduces the required computational cost of the existing face recognition models without the need for designing a particular architecture or accessing real training data.  
QuantFace introduces privacy-friendly synthetic face data to the quantization process to mitigate potential privacy concerns and issues related to the accessibility to real training data.
Through extensive evaluation experiments on seven benchmarks and four network architectures, we demonstrate that QuantFace can successfully reduce the model size up to 5x while maintaining, to a large degree, the verification performance of the full-precision model without accessing real training datasets. All training codes are publicly available \footnote{\url{https://github.com/fdbtrs/QuantFace}}.
\end{abstract}

%
\IEEEpeerreviewmaketitle

\section{Introduction}
\label{sec:intro}
Recent high performing deep neural networks (DNN) rely on over-parameterized networks with high computational cost \cite{resnet,senet}. State-of-the-art (SOTA) face recognition (FR) models followed this common trend by relying on over-parameterized DNN \cite{deng2019arcface,elasticface,DBLP:conf/fgr/CaoSXPZ18}.
However, deploying such extremely large models with hundreds of megabytes (MB) of memory requirements on embedded \textcolor{black}{devices and} other use-cases constrained by the computational capabilities and high throughput \textcolor{black}{requirements is} still challenging \cite{martinez2021benchmarking,DBLP:conf/iccvw/DengGZDLS19}.

Enabling FR on domains limited with computational capabilities requires designing a special architecture or compressing the current solutions to meet the computational requirements of the deployment environments.
Several efficient FR models have been proposed in the literature \cite{mobilefacenet,shufflefacenet,pocketnet,vargfacenet}. The core idea of many of these works depended on utilizing  efficient architecture designed for common computer version tasks such as MixNets, \cite{mixnet}, MobileNetV2 \cite{mobilenetv2}, ShuffleNet  \cite{shufflenetv2}, VarGNet \cite{vargnet} for FR \cite{mixfacenet,mobilefacenet,shufflefacenet,vargfacenet}. 
Very recently, few of works \cite{pocketnet,DBLP:conf/aaai/Wang21} proposed architectures based on face-specific neural architecture search (NAS) for efficient FR.
However, none of all previous efficient FR works \cite{mobilefacenet,shufflefacenet,mixfacenet,vargfacenet,pocketnet,DBLP:conf/aaai/Wang21} explored the potential of using model quantization to reduce the computational cost of existing widely used FR architectures, e.g. ResNet \cite{resnet,deng2019arcface} or SE-ResNet \cite{senet,DBLP:conf/fgr/CaoSXPZ18}. 

Model quantization approaches compress the DNN by reducing the number of bits required to represent each weight, e.g., using \textcolor{black}{a} lower precision format than full-precision (FP) floating-point, such as 4-bit \cite{banner2018post, yao2021hawq} or 8-bit \cite{krishnamoorthi2018quantizing, jacob2018quantization} signed integer.  
Such methods have shown great success in reducing the computation cost of DNN and are supported by most deep learning accelerators \cite{NEURIPS2019_9015,DBLP:conf/osdi/AbadiBCCDDDGIIK16}.
Model quantization enables performance gains in different areas. 
First, it reduces the model size, which can be directly measured using the number of bits required to represent each parameter. For example, applying model quantization using 8bit bandwidth on ResNet100 \cite{resnet,deng2019arcface}  (65.2M parameters) reduces the  model size from 261.2MB to 65.3 MB.
Second, many deep learning accelerators such as Pytorch \cite{NEURIPS2019_9015} and TensorFlow \cite{DBLP:conf/osdi/AbadiBCCDDDGIIK16} can run a quantized model faster than a FP one. For example, Pytorch \cite{NEURIPS2019_9015} can run a quantized model 2-4x faster than the FP model and reduce the required memory bandwidth by 2-4x \cite{NEURIPS2019_9015,krishnamoorthi2018quantizing}. However, the exact inference speed and memory bandwidth are highly dependent on the underlying \textcolor{black}{hardware and} deep learning accelerator \cite{NEURIPS2019_9015}. 
Once the model is quantized, the model weights and quantization parameters need to be tuned and calibrated.
These processes commonly require access to the training data, either entirely or partially \cite{choukroun2019low}. 

\begin{figure*}[t!]
     \centering
         \includegraphics[width=0.80\textwidth]{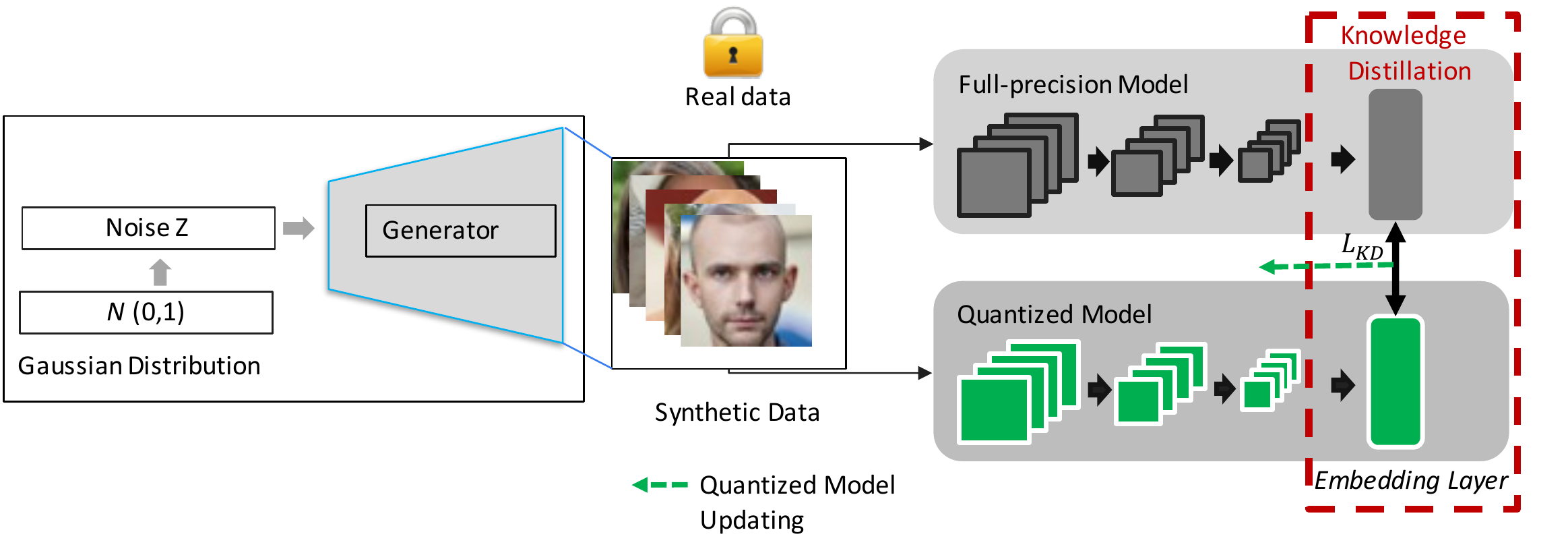}
         \vspace{-2mm}
         \caption{ An overview of the proposed QuantFace framework. Given a Gaussian noise Z from a normal distribution, the pretrained generator produces a fake data sample and feeds it into the FP and the b-bit quantized model. The KD loss is then computed between the normalized feature embeddings of the FP model and the quantized model.     }
 \label{fig:framewrok}
 \vspace{-5mm}
\end{figure*}

This need for data if quantization is used on FR networks follows that of deep FR models reliance on large-scale training datasets \cite{mobilefacenet,mixfacenet,deng2019arcface} such as MS1M \cite{guo2016ms}, VGGFace2 \cite{DBLP:conf/fgr/CaoSXPZ18} and CASIA-WebFace \cite{DBLP:journals/corr/YiLLL14a}. 
Existing efficient FR solutions are not different as they also require face image databases, whether for conventional training and/or knowledge distillation (KD) from teacher networks \cite{mobilefacenet,shufflefacenet,mixfacenet,vargfacenet,pocketnet,DBLP:conf/aaai/Wang21}.
Most of the recent face datasets used in the literature have been collected from the web \cite{DBLP:journals/corr/abs-2102-00813}.
According to \cite{DBLP:journals/corr/abs-2102-00813}, 45 of face datasets have been created after 2014, and around 78\% of these datasets are derived from the web.
However, it is not a trivial task, and it may not be feasible to further collect face datasets for biometric processing from the web due to legal privacy issues in some countries \cite{sface,Qiu_2021_ICCV,gdpr,Damer_2022_CVPR}.
Privacy regulations, just as the GDPR \cite{gdpr}, gives individuals the right to withdraw their consent to store or use their private data \cite{DBLP:journals/tifs/MedenRTDKSRPS21,gdpr}, a process that can practically be very challenging when a database is widely distributed, which puts the privacy rights of individuals in jeopardy.
Following such concerns, datasets such as VGGFace2 \cite{DBLP:conf/fgr/CaoSXPZ18} and CASIA-WebFace \cite{DBLP:journals/corr/YiLLL14a} are not anymore publicly accessible in many countries. Companies like Facebook announced that they will shut down their FR system due to such privacy concerns \cite{facebook}.
This motivated several recent works to explore the \textcolor{black}{ potential} of using synthetically generated face data \cite{sface,Qiu_2021_ICCV,DBLP:journals/access/RongZL20,Damer_2022_CVPR}.

This work presents contributions towards enhancing the compactness of FR models while maintaining high accuracy in a privacy-friendly process that does not require real face databases. 
Towards that, this work is, to the best of our knowledge, the first to regulate the FR computational cost by applying quantization-aware training.
We additionally propose a training paradigm involving KD that uses synthetically generated face data to fine-tune the quantized model \textcolor{black}{and adapt} the quantization operator parameters.
We empirically prove the success of the proposed approach in reducing the bit bandwidth up to 5x of the evaluated models while maintaining, to a large degree, the model verification performance. 
Additionally, the use of synthetic data within the proposed training paradigm proved to be highly effective and produced quantized models that performed competitively with models quantized using real data, even outperforming them in many experimental setups.
Our quantized models based on synthetic data did outperform other full precision models with higher memory footprints, as will be demonstrated later in this work.


\vspace{-2mm}
\section{Related works}
\vspace{-1mm}
There is a large body of works that proposed efficient FR models based on designing compact convolution building \textcolor{black}{blocks} \cite{mobilefacenet,shufflefacenet,mixfacenet,vargfacenet}.  
Such efficient FR models followed those of deep image classification \cite{mixnet,mobilenetv2,shufflenetv2}, which in terms evolved from the depthwise separable convolution \cite{xception}. 
Additionally, efficient FR models \cite{mobilefacenet,shufflefacenet,mixfacenet,vargfacenet} opted to replace the fully connected (FC) layer on the top of CNN with a global depthwise convolution to reduce the large number of parameters caused by the FC layer.
MobileFaceNets \cite{mobilefacenet} is a popular network architecture that has been widely adopted in different compact FR solutions \cite{airface,DBLP:conf/iccvw/DengGZDLS19}.
MobileFaceNets contains around one million trainable parameters with 443M FLOPs. 
MobileFaceNets architecture is based on the residual bottlenecks proposed by MobileNetV2 \cite{mobilenetv2} and depth-wise separable convolutions layer.
VarGFaceNet \cite{vargfacenet} deployed the variable group convolutional network proposed by VarGNet \cite{vargnet} to design a compact FR model with 5m trainable parameters. ShuffleFaceNet \cite{shufflefacenet} is a compact FR model based on ShuffleNet-V2 \cite{shufflenetv2}. 
MixFaceNets \cite{mixfacenet} proposed a family of efficient FR models by extending the MixConv block \cite{mixnet} with a channel shuffle operation \cite{shufflenetv2} aiming at increasing the discriminative ability of MixFaceNets.

An alternative to previous handcrafted DNN designs is utilizing NAS, KD, or a combination of different model compression techniques.
KD transfers the acquired knowledge learned by a larger network to a smaller one \cite{DBLP:journals/corr/HintonVD15,marco_FG}. KD has shown great success in improving the verification performance of compact FR models \cite{pocketnet,vargfacenet,s22051921}.
Furthermore, the combination of KD with other model compression techniques such as compact model design \cite{vargfacenet}, or NAS \cite{DBLP:conf/aaai/Wang21,pocketnet} \textcolor{black}{ demonstrated} very promising accuracies in FR. 

One must note that all the discussed FR models are built with FP single floating-point, and none has adopted model quantization techniques.
Additionally, all these works required the privacy-sensitive use of real face data, either during their conventional training or during the KD from a larger pre-trained network. This stresses the main contributions of this work, i.e., the use of unlabeled synthetic data in the proposed quantization paradigm.

\vspace{-1mm}
\section{Methodology}
\label{sec:methodogy}
\vspace{-1mm}
This work presents a privacy-friendly framework to minimize the computational complexity of deep learning-based FR models.
Towards that, given an FR model with a FP floating point, we propose to quantize the weights and activations to b-bit precision format through uniform quantization aware training.   
The quantization process commonly requires fine-tuning/retraining the quantized model to adjust the quantization operator parameters and recover the model accuracy after applying the quantization process.  
Quantization usually requires the original model training data, or part of it, for this calibration or fine-tuning \cite{DBLP:conf/cvpr/JacobKCZTHAK18}.
This original data may not be accessible after training the model due to restrictions related to privacy, security, data ownership, and data availability restrictions (e.g., model shared from a data owner to a third party). 
To mitigate this challenge and promote privacy-aware solutions, our proposed framework utilizes synthetically generated face data to fine-tune and calibrate the quantized model.
Moreover, to maintain the performance and enable the use of unlabeled synthetic data, the proposed framework combines the quantization process with KD during the fine-tuning phase. This step enables fine-tuning the quantized model without accessing the real training dataset or any prior assumption about the training dataset labels.
Figure \ref{fig:framewrok} presents an overview of the proposed framework.

This section presents first the quantization process applied to the FP floating-point FR model. Then, it presents the training paradigm utilized to fine-tune the quantized model.

\vspace{-1mm}
\subsection{Model Quantization}
\vspace{-1mm}
Quantization involves two main operators: Quantize and Dequantize.   
Let $x \in [\beta,\alpha]$ be a real value and $b$ is the bit width of a low precision format.  
$\beta$ and $\alpha$ are minimum and maximum values of $x$.
A $2^b$ possible integer values can be represented using  b-bit format and the value range of a b-bit signed integer precision format is between \textcolor{black}{ $[-2^{b-1},2^{b-1}-1]$}. 
A real value of a FP model in this work refers to a 32-bit single-precision floating-point.
Quantization maps a real value $x \in [\beta,\alpha]$ to lie within the range value of low precision b-bit \textcolor{black}{$[-2^{b-1},2^{b-1}-1]$}. 
Quantization operator consists of two processes: value transformation and clipping process. Formally, the transformation process that maps $x$ into $x_q$ can be defined as follows:
\vspace{-1mm}
\begin{equation}
\label{trans}
    T(x, s, z) = round(\frac{x}{s} - z),
\end{equation}
where $round()$ is the rounding method that defines the rounding step in which a value is rounded up or down to an integer. $z$ is a constant parameter (zero-point) of the same type as the quantized value. 
It represents the quantized value $x_q$ corresponding to the real zero value. 
$s$ is a real-valued scaling factor that \textcolor{black}{divides} the real value range into a number of fractions. 
In asymmetric quantization ($-\alpha \neq \beta$), the scaling factor $s$ and zero-point are defined as follows \cite{DBLP:conf/cvpr/JacobKCZTHAK18}:
\vspace{-1mm}
\begin{equation}
    s= \frac{\alpha-\beta}{2^b-1}.
\end{equation}
\vspace{-1mm}
\begin{equation}
    z=round (\beta . \frac{2^b-1}{\alpha-\beta} + 2^{b-1}).
\end{equation}
The clipping process clips $x_t$ (the output of Equation \ref{trans}) to lie within the range \textcolor{black}{$[-2^{b-1},2^{b-1}-1]$}. The clip operation can be defined as follows:
\vspace{-1mm}
\begin{equation}
    clip(x_t,l , u)= 
    \begin{cases}
    l &\text{if }  x_t < l \\
    x &\text{if }  l \leq x_t \leq u \\
    u &\text{if }  x > u,
    \end{cases}
\end{equation}
\vspace{-1mm}
where $l$ and $u$ are the minimum and maximum values of the quantization value range. 
The quantization of real value $x$ to a b-bit precision format is given by:
\vspace{-1mm}
\begin{equation}
    Q(x, b)= clip(T(x, s, z), -2^{b-1}, 2^{b-1}-1 ).
\end{equation}
Dequantization operation that approximates the real value of the quantized one is defined as follows:
\vspace{-1mm}
\begin{equation}
    D(q) = s \cdot (q + z).
\end{equation}
\vspace{-1mm}
\paragraph{Tensor Quantization Granularity}
Tensor quantization granularity defines how the quantization operator parameters are calculated and shared among the model tensors \cite{DBLP:journals/corr/abs-2103-13630}.
Quantization granularity \cite{DBLP:journals/corr/abs-2103-13630} can be categorized into three groups: per-layer \cite{DBLP:journals/corr/abs-1806-08342}, per-group of channels \cite{DBLP:conf/aaai/ShenDYMYGMK20}  and per-channel \cite{DBLP:conf/fpga/0001WDGCLWKW21,DBLP:conf/cvpr/JacobKCZTHAK18,DBLP:journals/corr/abs-1806-08342,DBLP:conf/eccv/ZhangYYH18}. In this work, we opt to use the popular choice of per-channel granularity as it provides a high quantization resolution and it has repeatedly led to high accuracy \cite{DBLP:conf/fpga/0001WDGCLWKW21,DBLP:conf/cvpr/JacobKCZTHAK18,DBLP:journals/corr/abs-1806-08342,DBLP:conf/eccv/ZhangYYH18}.  
Per-channel granularity uses \textcolor{black}{a fixed quantization parameter for each channel}, independently from other channels.


\begin{table*}[!h]
\vspace{3mm}
\centering
\begin{tabular}{|l|l|l|l|l|lllll|ll|}
\hline
\multirow{2}{*}{Model} &
  \multirow{2}{*}{param} &
  \multirow{2}{*}{\begin{tabular}[c]{@{}l@{}}Quantization \\ data\end{tabular}} &
  \multirow{2}{*}{Bits} &
  \multirow{2}{*}{Size (MB)} &
  \multicolumn{1}{l|}{LFW} &
  \multicolumn{1}{l|}{CFP} &
  \multicolumn{1}{l|}{AgeDb-30} &
  \multicolumn{1}{l|}{CALFW} &
  CPLFW &
  \multicolumn{1}{l|}{IJB-C} &
  IJB-B \\ \cline{6-12} 
 &
   &
   &
   &
   &
  \multicolumn{5}{l|}{Accuracy (\%)} &
  \multicolumn{2}{l|}{TAR@FAR 1e-4 (\%)} \\ \hline
\multirow{5}{*}{ResNet100} &
  \multirow{5}{*}{65.2M} &
  - &
  FP32 &
  261.22 &
  \multicolumn{1}{l|}{99.83} &
  \multicolumn{1}{l|}{98.40} &
  \multicolumn{1}{l|}{98.33} &
  \multicolumn{1}{l|}{96.13} &
  93.22 &
  \multicolumn{1}{l|}{96.50} &
  95.25 \\ \cline{3-12} 
 &
   &
  Real data &
  w8a8 &
  65.31 &
  \multicolumn{1}{l|}{\textbf{99.80}} &
  \multicolumn{1}{l|}{\textbf{98.31}} &
  \multicolumn{1}{l|}{\textbf{98.13}} &
  \multicolumn{1}{l|}{\textbf{96.05}} &
  \textbf{92.92} &
  \multicolumn{1}{l|}{\textbf{96.38}} &
  \textbf{95.13} \\ \cline{3-12} 
 &
   &
  \textcolor{black}{Synthetic} data &
  w8a8 &
  65.31 &
  \multicolumn{1}{l|}{\textbf{99.80}} &
  \multicolumn{1}{l|}{98.14} &
  \multicolumn{1}{l|}{97.95} &
  \multicolumn{1}{l|}{96.02} &
  92.90 &
  \multicolumn{1}{l|}{96.09} &
  94.74 \\ \cline{3-12} 
 &
   &
  Real data &
  w6a6 &
  49.01 &
  \multicolumn{1}{l|}{\textbf{99.55}} &
  \multicolumn{1}{l|}{89.14} &
  \multicolumn{1}{l|}{95.85} &
  \multicolumn{1}{l|}{95.42} &
  85.63 &
  \multicolumn{1}{l|}{85.80} &
  84.08 \\ \cline{3-12} 
 &
   &
  \textcolor{black}{Synthetic} data &
  w6a6 &
  49.01 &
  \multicolumn{1}{l|}{99.45} &
  \multicolumn{1}{l|}{\textbf{91.00}} &
  \multicolumn{1}{l|}{\textbf{96.43}} &
  \multicolumn{1}{l|}{\textbf{95.58}} &
  \textbf{86.60} &
  \multicolumn{1}{l|}{\textbf{87.00}} &
  \textbf{85.06} \\ \hline
\multirow{5}{*}{ResNet50} &
  \multirow{5}{*}{43.6M} &
  - &
  FP32 &
  174.68 &
  \multicolumn{1}{l|}{99.80} &
  \multicolumn{1}{l|}{98.01} &
  \multicolumn{1}{l|}{98.08} &
  \multicolumn{1}{l|}{96.10} &
  92.43 &
  \multicolumn{1}{l|}{95.74} &
  94.19 \\ \cline{3-12} 
 &
   &
  Real data &
  w8a8 &
  43.67 &
  \multicolumn{1}{l|}{\textbf{99.78}} &
  \multicolumn{1}{l|}{\textbf{97.70}} &
  \multicolumn{1}{l|}{\textbf{98.00}} &
  \multicolumn{1}{l|}{\textbf{96.00}} &
  \textbf{92.17} &
  \multicolumn{1}{l|}{\textbf{95.66}} &
  \textbf{94.15} \\ \cline{3-12} 
 &
   &
  \textcolor{black}{Synthetic} data &
  w8a8 &
  43.67 &
  \multicolumn{1}{l|}{\textbf{99.78}} &
  \multicolumn{1}{l|}{97.43} &
  \multicolumn{1}{l|}{97.97} &
  \multicolumn{1}{l|}{95.87} &
  92.08 &
  \multicolumn{1}{l|}{95.18} &
  93.67 \\ \cline{3-12} 
 &
   &
  Real data &
  w6a6 &
  32.77 &
  \multicolumn{1}{l|}{\textbf{99.70}} &
  \multicolumn{1}{l|}{95.00} &
  \multicolumn{1}{l|}{97.17} &
  \multicolumn{1}{l|}{\textbf{95.87}} &
  90.17 &
  \multicolumn{1}{l|}{\textbf{91.74}} &
  \textbf{90.07} \\ \cline{3-12} 
 &
   &
  \textcolor{black}{Synthetic} data &
  w6a6 &
  32.77 &
  \multicolumn{1}{l|}{99.68} &
  \multicolumn{1}{l|}{\textbf{95.17}} &
  \multicolumn{1}{l|}{\textbf{97.43}} &
  \multicolumn{1}{l|}{95.70} &
  \textbf{90.38} &
  \multicolumn{1}{l|}{90.72} &
  89.44 \\ \hline
\multirow{5}{*}{ResNet18} &
  \multirow{5}{*}{24.0M} &
  - &
  FP32 &
  96.22 &
  \multicolumn{1}{l|}{99.67} &
  \multicolumn{1}{l|}{94.47} &
  \multicolumn{1}{l|}{97.13} &
  \multicolumn{1}{l|}{95.70} &
  89.73 &
  \multicolumn{1}{l|}{93.56} &
  91.64 \\ \cline{3-12} 
 &
   &
  Real data &
  w8a8 &
  24.10 &
  \multicolumn{1}{l|}{\textbf{99.63}} &
  \multicolumn{1}{l|}{\textbf{94.46}} &
  \multicolumn{1}{l|}{97.03} &
  \multicolumn{1}{l|}{\textbf{95.72}} &
  89.48 &
  \multicolumn{1}{l|}{\textbf{93.56}} &
  \textbf{91.57} \\ \cline{3-12} 
 &
   &
  \textcolor{black}{Synthetic} data &
  w8a8 &
  24.10 &
  \multicolumn{1}{l|}{99.55} &
  \multicolumn{1}{l|}{94.04} &
  \multicolumn{1}{l|}{\textbf{97.07}} &
  \multicolumn{1}{l|}{95.58} &
  \textbf{89.53} &
  \multicolumn{1}{l|}{92.87} &
  91.01 \\ \cline{3-12} 
 &
   &
  Real data &
  w6a6 &
  18.10 &
  \multicolumn{1}{l|}{99.52} &
  \multicolumn{1}{l|}{93.23} &
  \multicolumn{1}{l|}{96.55} &
  \multicolumn{1}{l|}{\textbf{95.58}} &
  88.37 &
  \multicolumn{1}{l|}{\textbf{93.03}} &
  \textbf{91.08} \\ \cline{3-12} 
 &
   &
  \textcolor{black}{Synthetic} data &
  w6a6 &
  18.10 &
  \multicolumn{1}{l|}{\textbf{99.55}} &
  \multicolumn{1}{l|}{\textbf{93.34}} &
  \multicolumn{1}{l|}{\textbf{96.62}} &
  \multicolumn{1}{l|}{95.32} &
  \textbf{89.05} &
  \multicolumn{1}{l|}{92.36} &
  90.38 \\ \hline
\multirow{5}{*}{MobileFaceNet} &
  \multirow{5}{*}{1.1M} &
  - &
  FP32 &
  4.21 &
  \multicolumn{1}{l|}{99.47} &
  \multicolumn{1}{l|}{91.59} &
  \multicolumn{1}{l|}{95.62} &
  \multicolumn{1}{l|}{95.15} &
  87.98 &
  \multicolumn{1}{l|}{90.88} & 88.54
   \\ \cline{3-12} 
 &
   &
  Real data &
  w8a8 &
  1.10 &
  \multicolumn{1}{l|}{\textbf{99.43}} &
  \multicolumn{1}{l|}{\textbf{91.40}} &
  \multicolumn{1}{l|}{\textbf{95.47}} &
  \multicolumn{1}{l|}{\textbf{95.05}} &
  \textbf{87.95} &
  \multicolumn{1}{l|}{\textbf{90.57}} &
  \textbf{88.32} \\ \cline{3-12} 
 &
   &
  \textcolor{black}{Synthetic} data &
  w8a8 &
  1.10 &
  \multicolumn{1}{l|}{99.35} &
  \multicolumn{1}{l|}{90.84} &
  \multicolumn{1}{l|}{94.37} &
  \multicolumn{1}{l|}{94.78} &
  87.73 &
  \multicolumn{1}{l|}{89.21} &
  86.98 \\ \cline{3-12} 
 &
   &
  Real data &
  w6a6 &
  0.79 &
  \multicolumn{1}{l|}{98.87} &
  \multicolumn{1}{l|}{\textbf{87.69}} &
  \multicolumn{1}{l|}{\textbf{93.03}} &
  \multicolumn{1}{l|}{93.30} &
  84.57 &
  \multicolumn{1}{l|}{\textbf{83.13}} &
  80.53 \\ \cline{3-12} 
 & 
   &
  \textcolor{black}{Synthetic} data &
  w6a6 &
  0.79 &
  \multicolumn{1}{l|}{\textbf{99.08}} &
  \multicolumn{1}{l|}{87.64} &
  \multicolumn{1}{l|}{91.77} &
  \multicolumn{1}{l|}{\textbf{93.48}} &
  \textbf{84.85} &
  \multicolumn{1}{l|}{82.94} &
  \textbf{80.58} \\ \hline
\end{tabular}%
\vspace{-1mm}
\caption{B-bit precision data vs. performance on LFW , CFP , AgeDb-30 , CALFW, CPLFW (Accuracy \%), IJB-C and IJB-B (TAR at FAR 1e-4). The results is reported for FP model (32-bit), quantized models to 8-bit (w8a8) and 6-bit (w6a6) using real and synthetic data. All decimal points are rounded up to two decimal places. The top verification performances under the same quantization settings (network architecture and bit bandwidth) are in bold.}
\label{tab:result}
\vspace{-5mm}
\end{table*}

\vspace{-1mm}
\subsection{Quantization-Aware Training (QAT)}
\vspace{-1mm}
QAT, utilized in QuantFace, is a common approach to adjust the quantized model parameters \cite{DBLP:conf/cvpr/WuLWHC16,DBLP:conf/cvpr/JacobKCZTHAK18}. This adjustment is often required for model quantization because rounding the weights of a pre-trained model often results in lower accuracy, especially if the weights and the activation functions have a wide range of values.
QAT inserts fake (simulated) quantization operations in the network to emulate inference-time quantization \cite{DBLP:conf/cvpr/JacobKCZTHAK18}. QAT requires training or fine-tuning the model. In QAT,  the forward and backward passes are usually carried out in a floating-point precision, and the model weights and activations are quantized after each gradient update \cite{DBLP:conf/cvpr/JacobKCZTHAK18}. 
After each training iteration, the derivatives of the quantized network weights need to be calculated to compute the loss gradients for backpropagation.
However, the gradients of the fake quantized operations are predominantly zero \cite{krishnamoorthi2018quantizing}, making the standard backpropagation not applicable. QuantFace follows the common QAT approaches \cite{krishnamoorthi2018quantizing,DBLP:conf/cvpr/JacobKCZTHAK18} by addressing this issue \textcolor{black}{by} using Straight Through Estimator (STE) \cite{DBLP:journals/corr/BengioLC13} to approximate the gradient of fake quantization operators using a threshold, where the derivatives of fake quantization operators are set to one for inputs within the clipping range, i.e. when $x$ is in the range $[\beta, \alpha]$.

\vspace{-1mm}
\subsection{Training paradigm}
\vspace{-1mm}
QAT requires access to the original labeled training dataset to fine-tune the quantized model \cite{DBLP:conf/cvpr/JacobKCZTHAK18}, which may be infeasible for privacy concerns \cite{DBLP:conf/cvpr/ChoiCEL20}. 
Very recently, a number of works proposed to fine-tune a quantized model with generated data from conditional generative models \cite{jacob2018quantization,DBLP:conf/eccv/XuLZLCLT20}. 
However, these works \cite{jacob2018quantization,DBLP:conf/eccv/XuLZLCLT20} required to generate data with labels that are aware of the data labels used to train the FP models. 
Such knowledge of training class labels is often unobtainable, given only a pre-trained FR model.
To solve this, we propose a solution that utilizes unlabeled synthetic data along with a KD-based training paradigm that wavers the requirement of training data labels. 
\textcolor{black}{
Unlike conventional KD \cite{DBLP:journals/corr/HintonVD15} that optimizes the classification output (require labeled data), the utilized KD-based training paradigm optimizes the feature representation needed for biometric verification i.e. given a batch of \textit{unlabeled} face images, the quantized model is fine-tuned to learn producing feature representation similar to the one learned by the full precision model 
as will be detailed in this section. }
We propose to use synthetically generated data from a Generative Adversarial Network (GAN) \cite{gan,DBLP:conf/nips/KarrasAHLLA20} to fine-tune the quantized model.
We sample noise $Z$ \textcolor{black}{from} a Gaussian distribution $N(0,1)$ and feed it into a pretrained generator $G$ to generate unlabeled synthetic data $x$, as shown in Figure \ref{fig:framewrok}. Formally, the synthetically generated data is obtained by:
\vspace{-1mm}
\begin{equation}
   x= G(z),  \quad  Z \sim N(0,1). 
\end{equation}
Then, the synthetically generated data is aligned and cropped (See Section \ref{sec:exp}).
This data is unlabeled, as the random $Z$ produces random identities unrelated to those used to train the FP model. Thus, the model cannot be directly fine-tuned with the generated data.
Given this restriction, we propose in this work to fine-tune the quantized model using KD from the FP model. Specifically, the quantized model is trained to learn feature embedding similar to the ones from the FP model in the normalized embedding space. 
\textcolor{black}{
During the fine-tuning phase, a batch of size $M$ of unlabeled synthetic images $X$ is sampled and fed into the quantized and full precision model to obtain feature embeddings $f^q$ and $f^t$, respectively.}
\textcolor{black}{
Then, $f^q$ and $f^t$ are normalized ($f^q=\frac{f^q}{||f^q||_2}$ ,$f^t=\frac{f^t}{||f^t||_2}$) and used to compute the $\mathcal{L}_{kd}$ loss based on the cosine distance between the normalized features.
Finally, the gradient of the $\mathcal{L}_{kd}$ loss function is computed and used to update the weight parameters. 
Different from supervised classification losses e.g. cross-entropy or conventional KD losses \cite{DBLP:journals/corr/HintonVD15} that require class label for calculating the loss value,  $\mathcal{L}_{kd}$ loss is calculated based on the feature embedding layers. Thus, it mitigates the need to have identity labels for the input training images.}
Formally, the $\mathcal{L}_{kd}$ loss is defined as \textcolor{black}{follows}:
\vspace{-2mm}
\begin{equation}
    \mathcal{L}_{kd}=1- \frac{1}{M}\sum\limits_{i \in M} \frac{f^q_i . f^t_i}{||f^q_i|| \;||f^t_i||}.
\end{equation}

\textcolor{black}{
Using synthetic data to train FR model might lead to sub-optimal verification performances \cite{Qiu_2021_ICCV} due to a possible domain gap between the real and synthetic data \cite{DBLP:conf/aaai/XuZNLWTZ20,DBLP:conf/cvpr/Sankaranarayanan18,DBLP:conf/cvpr/LeePYL20}, especially if the model is trained from scratch to learn identity representation \cite{Qiu_2021_ICCV}, which might require conducting domain adaption e.g. adversarial domain adaptation \cite{DBLP:conf/aaai/XuZNLWTZ20,Qiu_2021_ICCV} to reduce such effect.
However, our goal in this work is not to learn identity representation from scratch through optimizing classification loss e.g. cross entropy, we rather fine-tune the pretrained model with synthetic data to adjust the model weights and quantization parameters after applying quantization process. 
Thus, by utilizing the proposed training paradigm, we restrict our use of the synthetic data to ensure that the response of the pretrained quantized model to input $x \in X$ during the fine-tuning phase is similar to the response of the full precision model to the same input $x \in X$.
We demonstrate that this process is not largely effected by the potential domain gap between the real and synthetic data by (1) comparing the model responses to real and synthetic data i.e. activation function value ranges of two models fine-tuned with real and synthetic data \ref{fig:act}, respectively and by (2) comparing the evaluation results of these models on real data benchmarks (details in Section \ref{sec:data_source}). 
}

\vspace{-1mm}
\section{Experimental setup}
\label{sec:exp}
\vspace{-1mm}
This section presents the baseline models with implementation details, model quantization implementation details, and evaluation benchmarks used in this work.

\vspace{-1mm}
\subsection{Baselines}
\label{sec:exp_baseline}
\vspace{-1mm}
The FP models in this work are trained with ArcFace loss \cite{deng2019arcface} on MS1MV2 dataset \cite{guo2016ms,deng2019arcface}. 
The MS1MV2 is a refined version of the MS-Celeb-1M \cite{guo2016ms} by \cite{deng2019arcface} containing 5.8M images of 85K identities. 
The baseline backbones are ResNet100 \cite{resnet,deng2019arcface}, ResNet50 \cite{resnet,deng2019arcface}, ResNet18 \cite{resnet,deng2019arcface}, and MobileFaceNet \cite{mobilefacenet}. 
We follow the training setting of \cite{deng2019arcface} to set the scale parameter $s$ to 64 and the margin $m$ to 0.5. 
We set the mini-batch size to 512 and trained the presented models on one Linux machine (Ubuntu 20.04.2 LTS) with Intel(R) Xeon(R) Gold 5218 CPU 2.30GHz, 512G RAM, and four Nvidia GeForce RTX-6000 GPUs. 
The FP models and the quantization operators in this paper are implemented using Pytorch \cite{NEURIPS2019_9015}.
All models are trained with Stochastic Gradient Descent (SGD) optimizer.
We used random horizontal flipping with a probability of 0.5 for data augmentation during the training.
We set the momentum to 0.9 and the weight decay to 5e-4. 
All the images in the evaluation and training datasets are aligned and cropped to $112 \times 112$, as described in \cite{deng2019arcface}.
The initial learning rate of the FP models is 0.1, and it is divided by 10 at 100K and 160K training iterations, following \cite{deng2019arcface}. The training is stopped after 180K iterations.

\subsection{Quantization implementation details}
We quantize the weights and activations of all the baseline FP models to two, 6-bit and 8-bit, precision formats. 
We reported the results of the quantized models under two settings. 
First, the quantized models are fine-tuned and calibrated with the original (real) training data, MS1MV2 \cite{guo2016ms,deng2019arcface} (described in Section \ref{sec:exp_baseline}). 
Second, the quantized models are fine-tuned and calibrated with the synthetically generated data.
We utilized the official open source implementation \footnote{\url{https://github.com/NVlabs/stylegan2-ada}} of StyleGAN2-ADA 
to randomly generate 0.5M synthetic face images.
These images are then cropped and aligned using the method described in Section \ref{sec:exp_baseline}.
In both settings, the quantized models are fine-tuned for 11K iterations with a learning rate of 1e-4.

\begin{figure}[!t]
     \centering
     \begin{subfigure}[b]{0.40\linewidth}
         \centering
         \includegraphics[width=\textwidth]{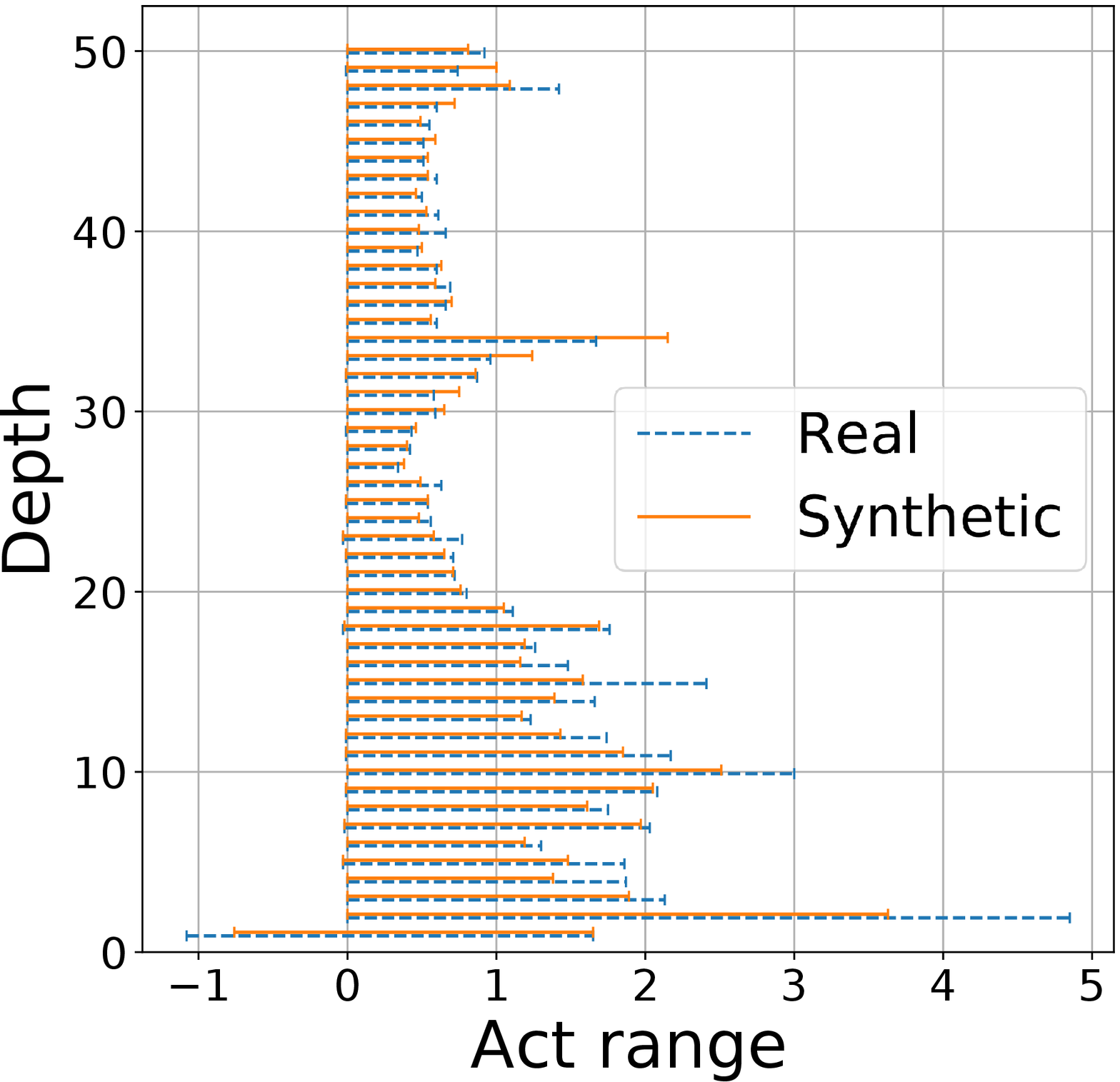}
         \caption{ResNet100 6-bit}
         \label{fig:r100_6}
     \end{subfigure}
      \begin{subfigure}[b]{0.40\linewidth}
         \centering
         \includegraphics[width=\textwidth]{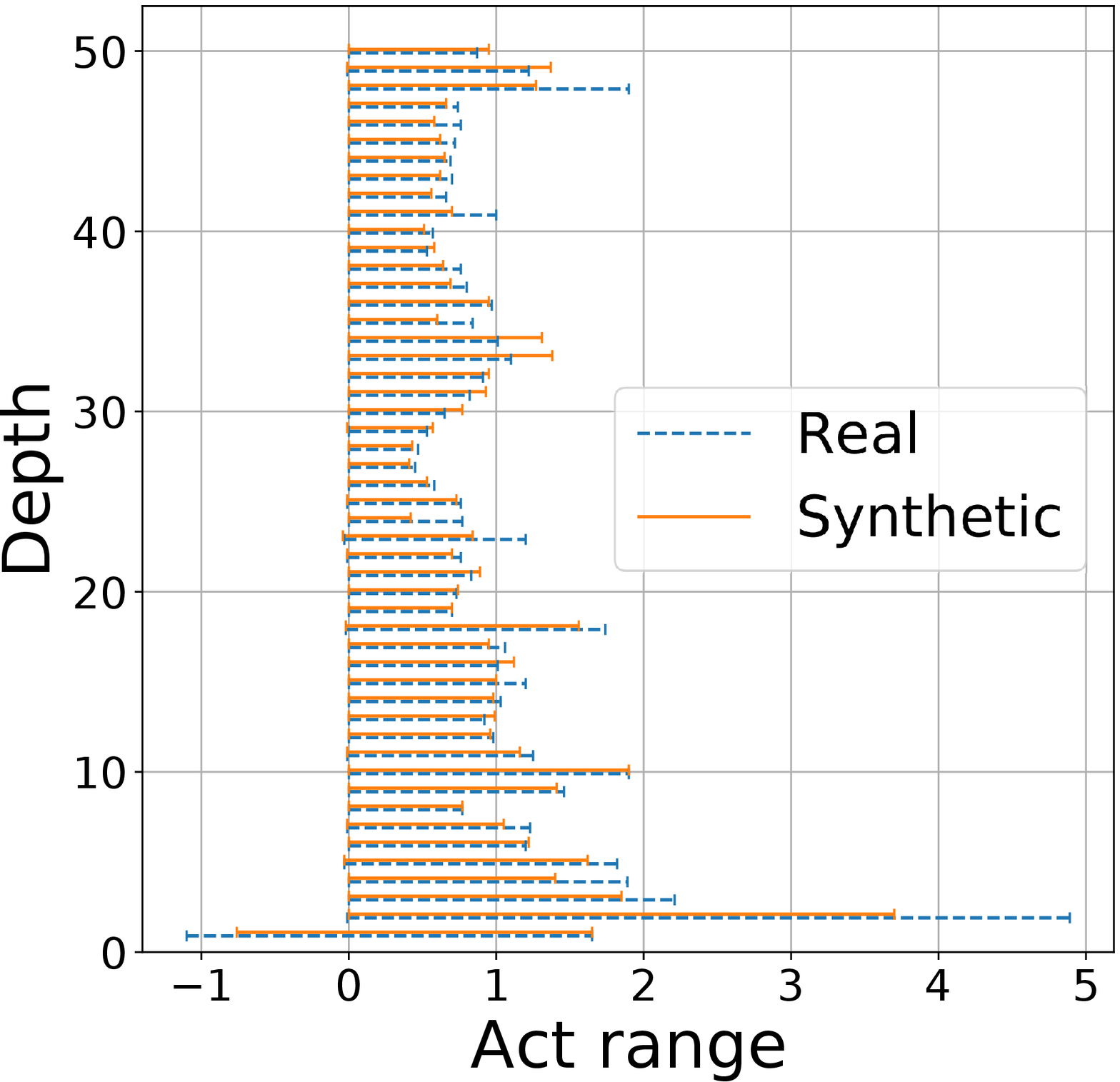}
         \caption{ResNet100 8-bit}
         \label{fig:r100_8}
     \end{subfigure}
     \vspace{-1mm}
             \caption{Correlation between the activation functions (Act) value ranges ($[\beta,\alpha]$) of the ResNet100 quantized using real (solid orange) and synthetic data (dashed black).
             The y-axis represents the depth of the backbone activation function e.g. depth 1 is the first activation function.  
             These plots represent a fixed-precision quantization bit bandwidth of 6-bit (\ref{fig:r100_6}) and 8-bit (\ref{fig:r100_8}).
             Each line in the plot represents the range value of the activation function where the start point is $\beta$ and end point is $\alpha$.
             The high correlation indicates that the quantized model is able to capture sufficient data information from the synthetic data, in comparison to real data.
             }
        \label{fig:act}
        \vspace{-4mm}
\end{figure}

\vspace{-1mm}
\subsection{Evaluation benchmarks and metrics} 
\vspace{-1mm}
The evaluation results of the FP and quantized models are reported on seven mainstream benchmarks: Labeled Faces in the Wild (LFW) \cite{LFWTech}, AgeDB-30 \cite{agedb}, Celebrities in Frontal-Profile in the Wild (CFP-FP) \cite{cfp-fp}, Cross-age LFW (CALFW) \cite{CALFW},  Cross-Pose LFW (CPLFW)  \cite{CPLFWTech}, IARPA Janus Benchmark–C and B (IJB-C) \cite{ijbc} and (IJB-B) \cite{DBLP:conf/cvpr/WhitelamTBMAMKJ17}. 
We follow the evaluation metrics defined in the utilized benchmarks as follows: LFW (accuracy), CA-LFW (accuracy), CP-LFW (accuracy), CFP-FP (accuracy), AgeDB-30 (accuracy),  IJB-C, and IJB-B (true acceptance rate at a false acceptance rate of 1e-4, noted as TAR at FAR1e-4).

\begin{figure*}[ht!]
     \centering
     \begin{subfigure}[b]{0.22\textwidth}
         \centering
         \includegraphics[width=\textwidth]{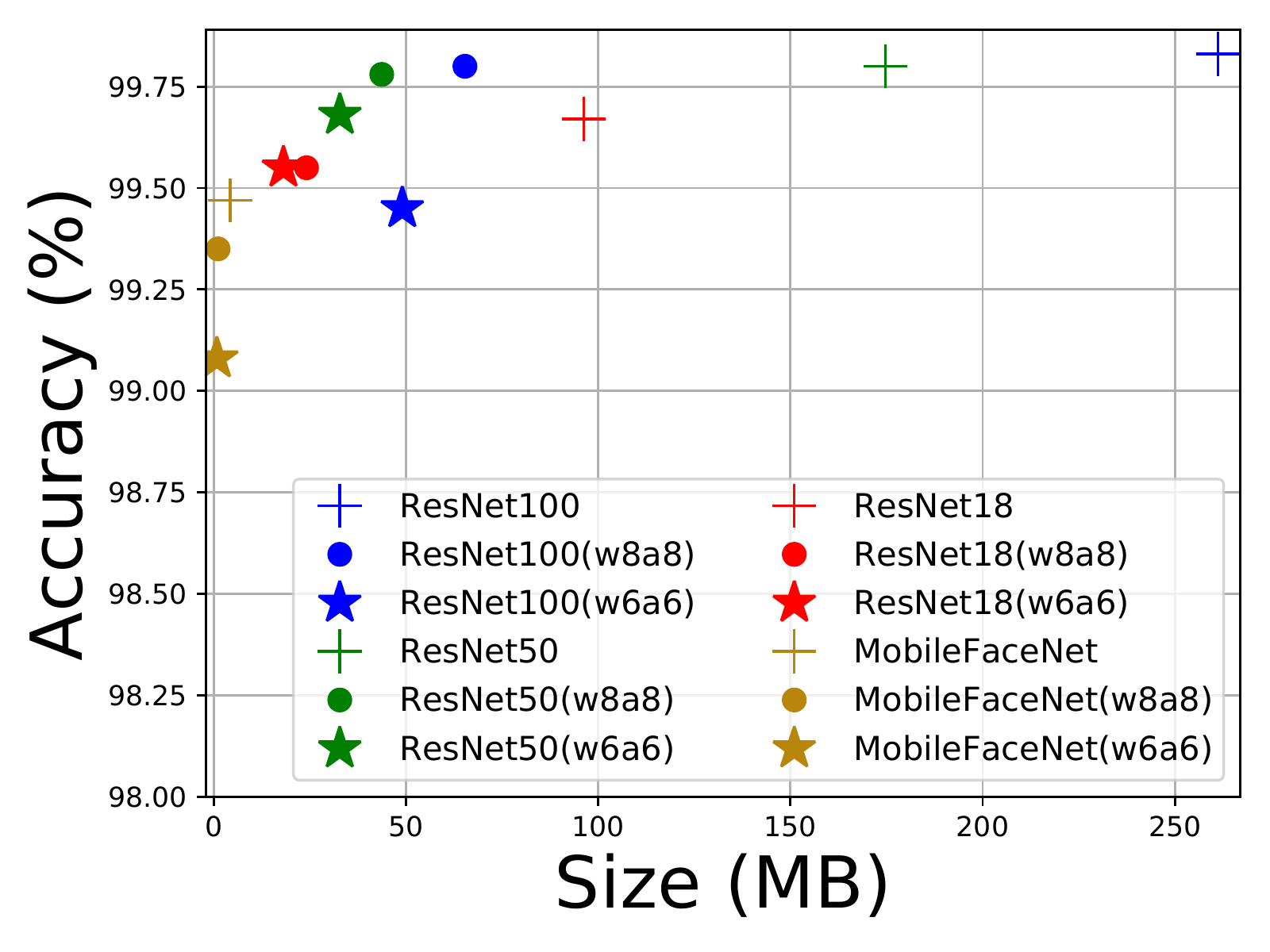}
         \caption{LFW}
         \label{fig:lfw}
     \end{subfigure}
      \begin{subfigure}[b]{0.22\textwidth}
         \centering
         \includegraphics[width=\textwidth]{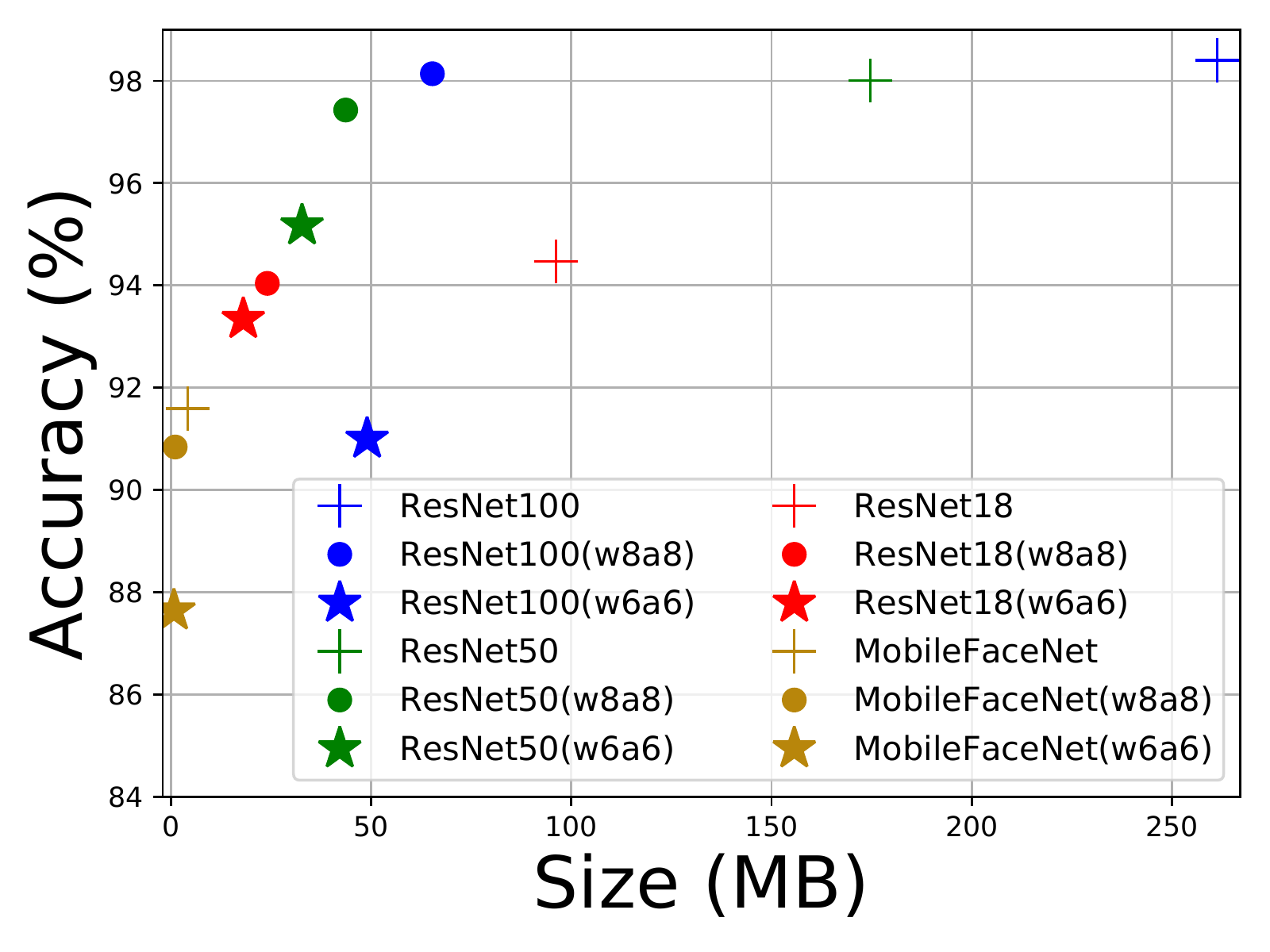}
         \caption{cfp}
         \label{fig:cfp}
     \end{subfigure}
           \begin{subfigure}[b]{0.22\textwidth}
         \centering
         \includegraphics[width=\textwidth]{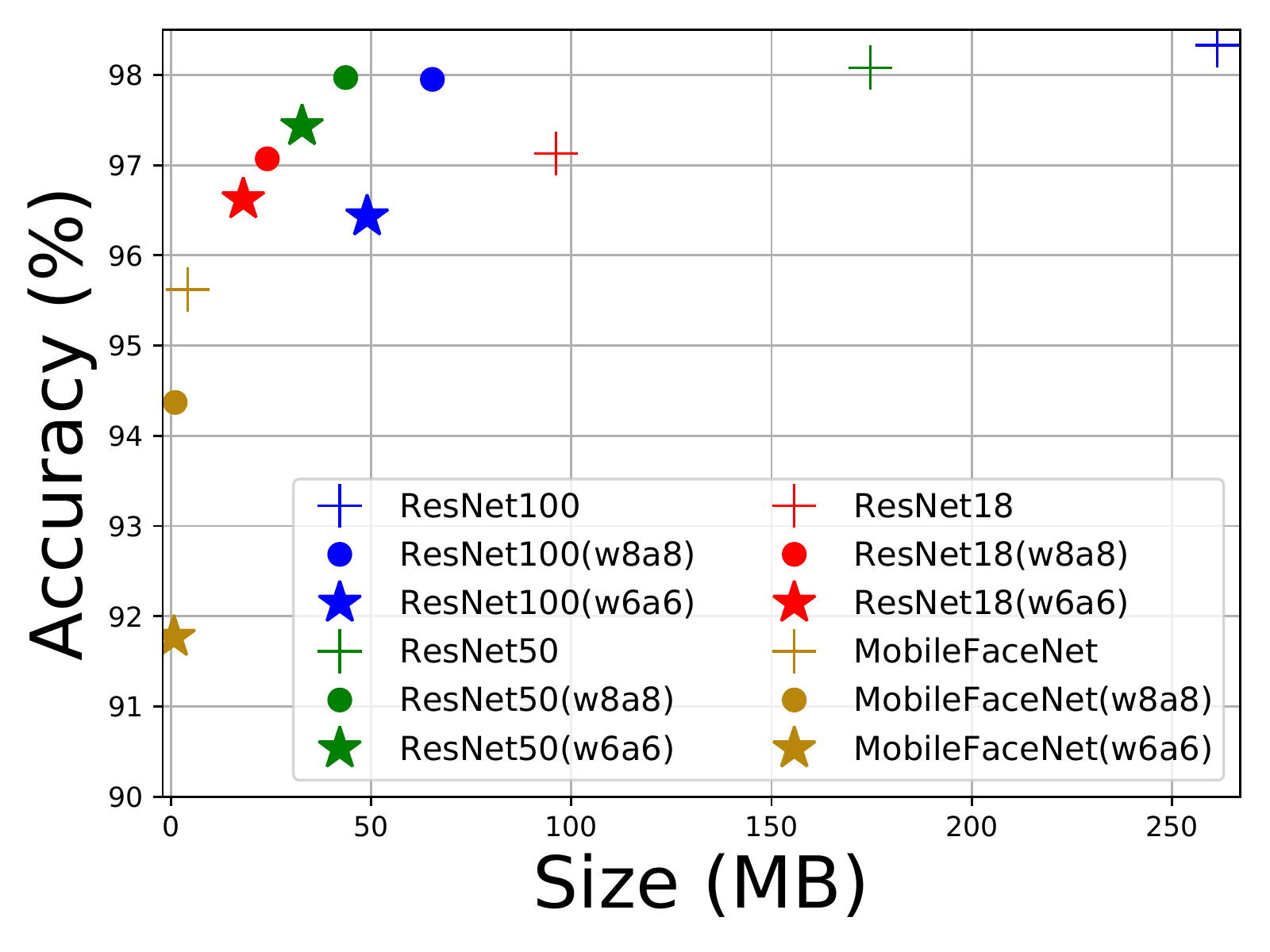}
         \caption{AgeDb-30}
         \label{fig:agedb}
     \end{subfigure}
    \begin{subfigure}[b]{0.22\textwidth}
         \centering
         \includegraphics[width=\textwidth]{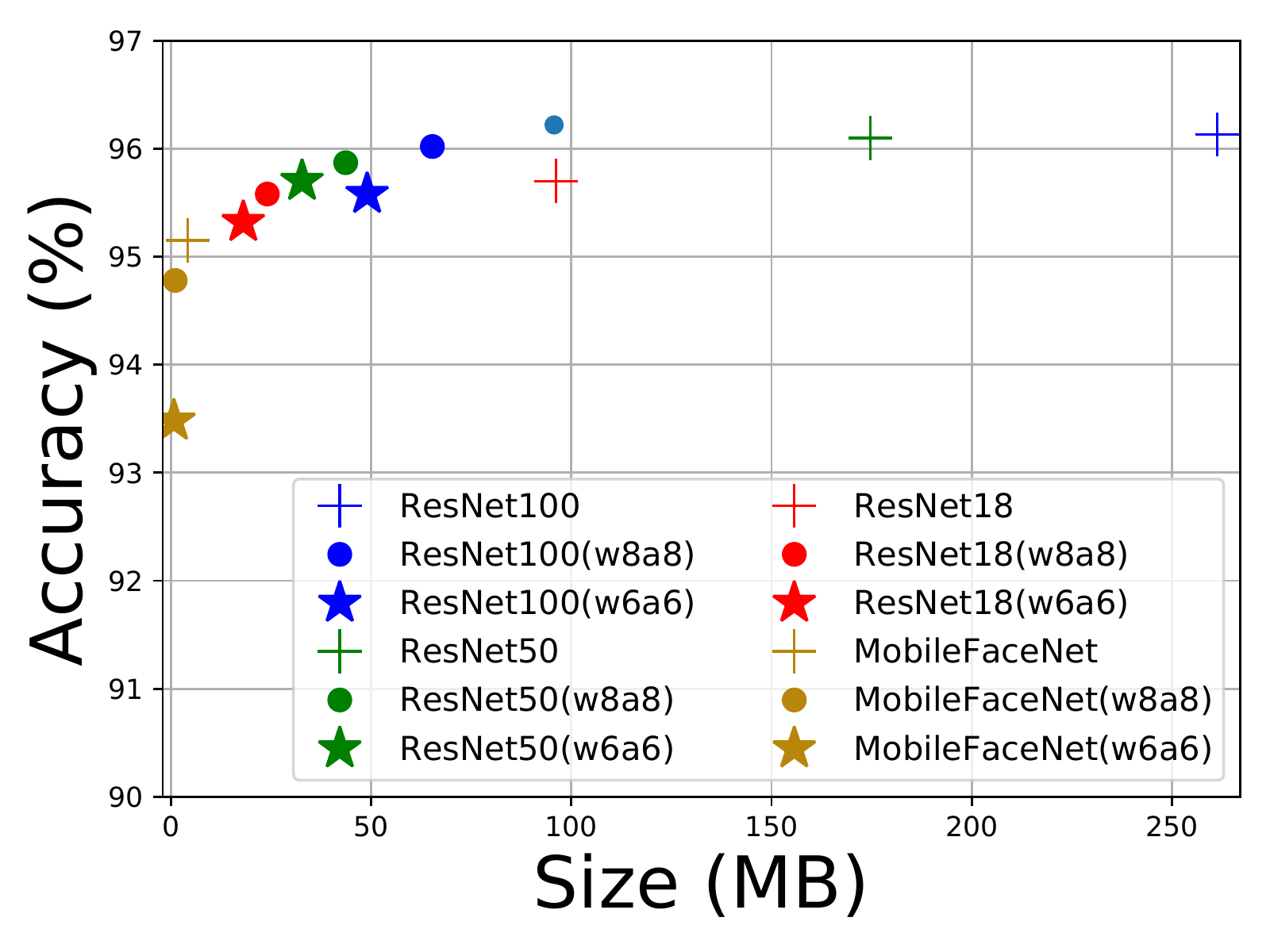}
         \caption{CALFW}
         \label{fig:calfw}
     \end{subfigure}
     
         \begin{subfigure}[b]{0.22\textwidth}
         \centering
         \includegraphics[width=\textwidth]{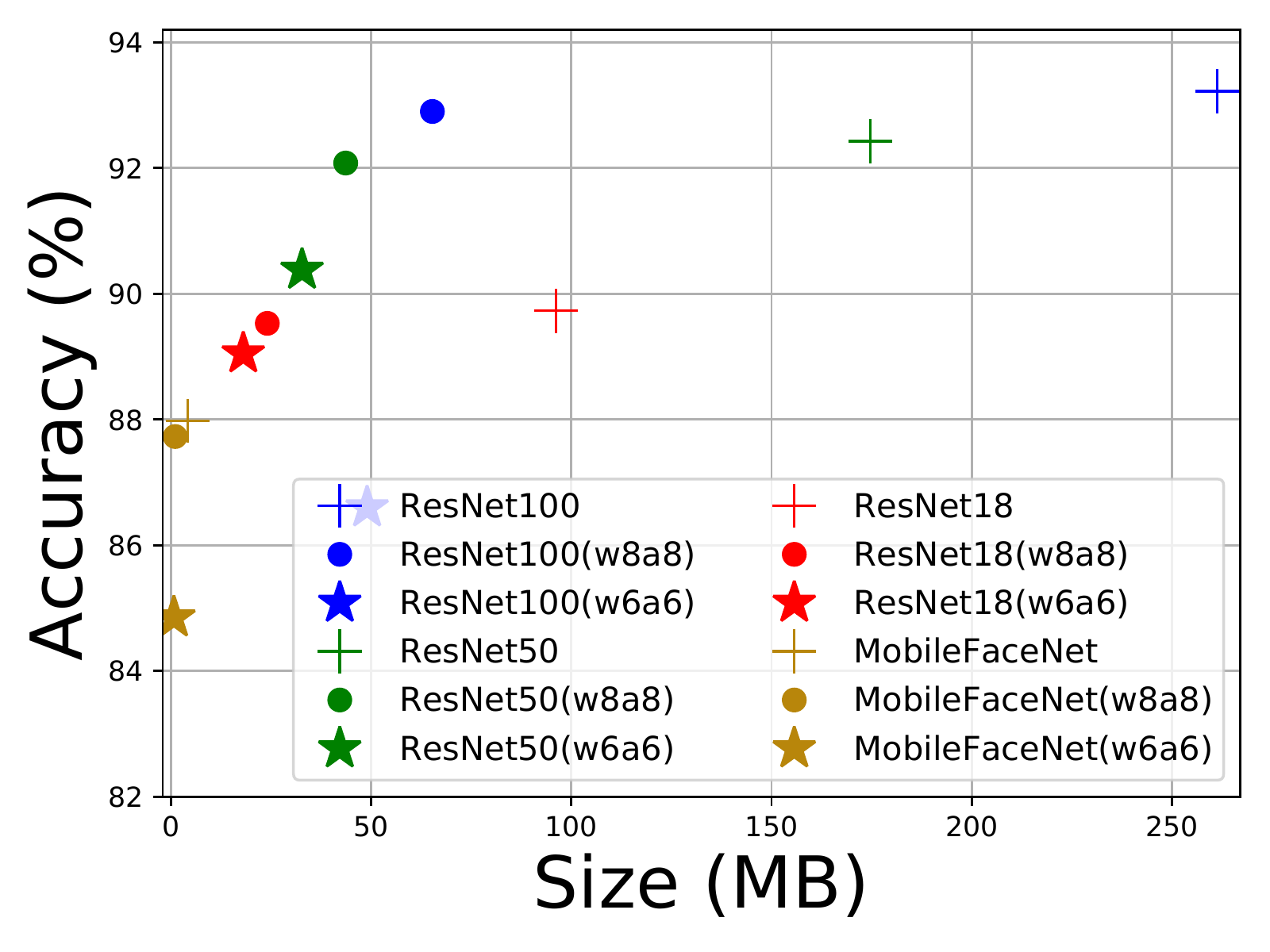}
         \caption{CPLFW}
         \label{fig:cplfw}
     \end{subfigure}
      \begin{subfigure}[b]{0.22\textwidth}
         \centering
         \includegraphics[width=\textwidth]{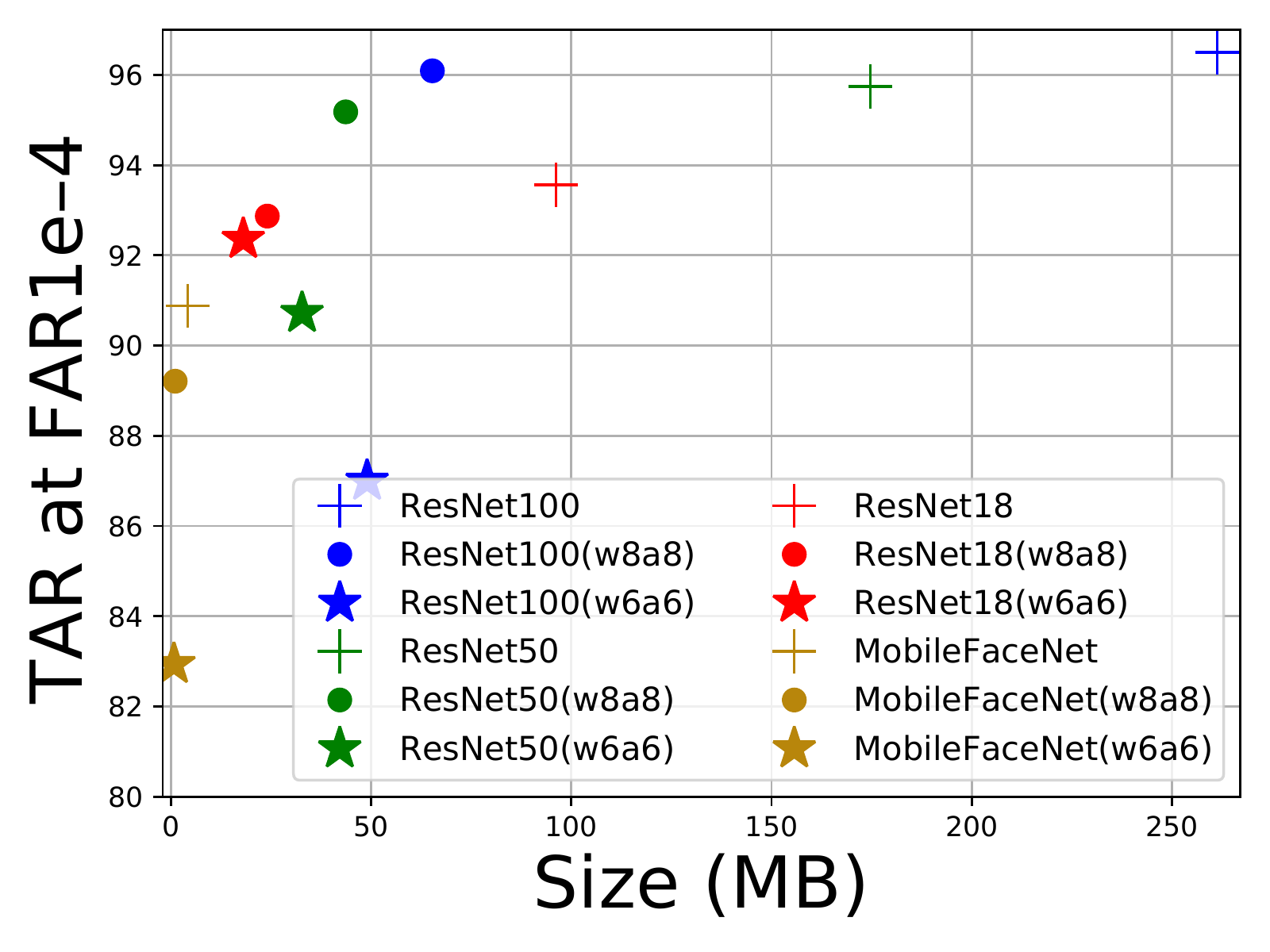}
         \caption{IJB-C}
         \label{fig:ijbc}
     \end{subfigure}
            \begin{subfigure}[b]{0.22\textwidth}
         \centering
         \includegraphics[width=\textwidth]{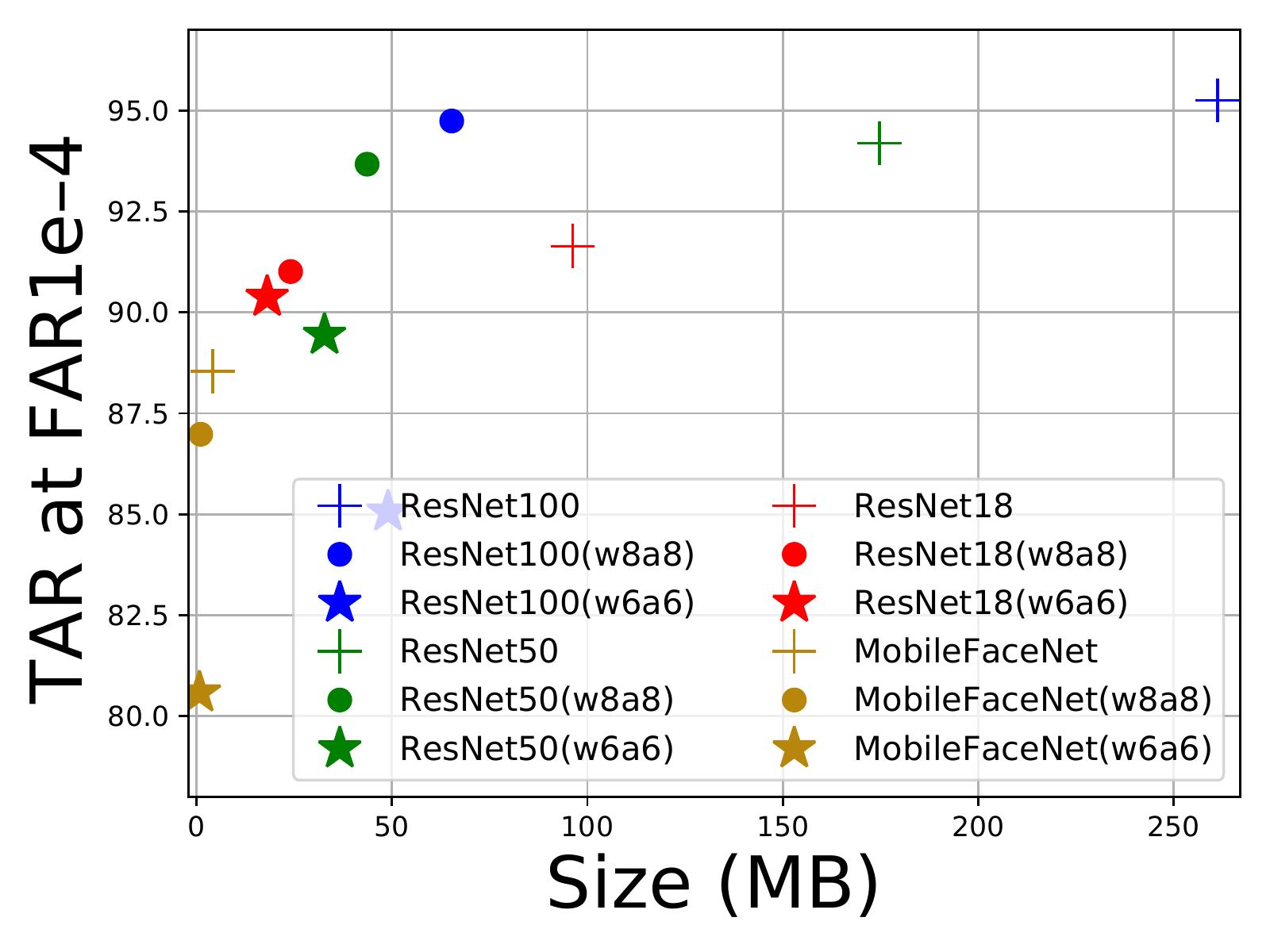}
         \caption{IJB-B}
         \label{fig:ijbb}
     \end{subfigure}
     \vspace{-2mm}
        \caption{The model size (in MB) vs. performance on LFW (accuracy), CFP (accuracy), AgeDB-30 (accuracy), CALFW (accuracy), CPLFW (accuracy), IJB-C (TAR at FAR1e-4), and  IJB-B (TAR at FAR1e-4). The FP models are marked with \textcolor{black}{ a cross}. The 8bit and 6bit quantized models using synthetic data are marked with full circles and stars, respectively. The models \textcolor{black}{, after quantization, only lose} marginal performance while their sizes are significantly reduced. Quantized models (e.g. ResNet100(w8a8), blue circle) in most cases outperforms larger full precision models (e.g. ResNet50 (green cross) and ResNet18 (red cross)).
       }
       \label{fig:tradeoff}
\vspace{-6mm}
\end{figure*}

\vspace{-1mm}
\section{Results}
\label{sec:result}
\vspace{-1mm}
Table \ref{tab:result} presents the achieved FR performance results by the FP models (ResNet100, ResNet50, ResNet18, and MobileFaceNet), along with the achieved ones by the quantized models to 8-bit weights and 8-bit activation (noted as w8a8), and to  6-bit weights and 6-bit activation (noted as w6a6) using synthetic or real data quantization. The results are grouped by each network architecture. In each group of rows, the results are first presented for the FP model (baseline), followed by the quantized models.  
The size (in MB) of the FP models is approximately 4x the number of parameters, i.e., each parameter requires 4 bytes.
In both real and synthetic data quantization settings, the reductions in bit bandwidth, and thus the model size, using w8a8 and w6a6, are around 4x and 5.3x, respectively. Also, model quantization enables performance gains in inference speed and memory bandwidth. However, the exact measures of inference speed and memory bandwidth depend on the underlying hardware and deep learning accelerator, as we discussed in Section \ref{sec:intro}.   Therefore, the presented results in this section are discussed as a trade-off between FR performance and the bit bandwidth, and thus, model size.
Figure \ref{fig:tradeoff} presents the trade-off between the model size (in MB) and the achieved verification performance by the FP floating-point 32 models (FP32) and their respective quantized models using synthetically generated data. 
The model that has the best trade-off between the verification performance and model size tends to be on the top left in the plot \ref{fig:tradeoff}.
The following observations can be made based on the achieved results in Table \ref{tab:result}:

\vspace{-1mm}
\subsection{Impact of 8-bit bandwidth quantization}
\vspace{-1mm}
When the model is quantized to 8-bit (w8a8 setting), the achieved verification performances in all experimental settings are slightly degraded. However, the bit bandwidth is significantly reduced (around 4x) when the considered models are quantized. For example, the achieved accuracy by ResNet100 (261.22 MB) on \textcolor{black}{AgeDb-30} is 98.33\%.
This accuracy slightly dropped to 98.13\% and 97.95\% when the ResNet-100 is quantized to 8-bit (65.31 MB) and fine-tuned with real and synthetic data, respectively. Similar observations can be made when all considered models are quantized to 8-bit. Another important observation can be drawn from the achieved results: when ResNet100 is quantized to 8-bit and fine-tuned with synthetic data (65.31 MB), it significantly outperformed ResNet18 (96.22 MB) on all considered benchmarks, resulting in around 30\% less model size. Impressively, the ResNet100 quantized to 8-bit outperformed the FP ResNet50 on most benchmarks while being more than 60\% smaller. 
This observation can be visually seen
by comparing the (x,y) positions of blue circle (ResNet100(w8a8)) and green cross (ResNet50) marks in
the trade-off plots of Figure \ref{fig:tradeoff}.
On large-scale evaluation \textcolor{black}{benchmarks}, the achieved TAR at FAR 1e-4 on IJB-C by quantized models using real and synthetic data is very competitive to the FP model. For example, the achieved verification performance by the FP ResNet19 is 93.56\% TAR at FAR 1e-4, and the achieved performance by the quantized models with real and synthetic data are 93.56\% and 92.87\%, respectively. 

\subsection{Impact of 6-bit bandwidth quantization}
Using 6-bit bandwidth, the achieved results by quantized models are, as expected, lower than the ones achieved by the 8-bit quantization. However, the reduction in model size is significantly higher than the 8-bit quantization. However, using 6-bit bandwidth can still achieve competitive results to the FP model for use-cases that are extremely limited with computational cost. 
Moving below 6-bit bandwidth, e.g., 4-bit, our experiments showed that none of the considered models converged during the fine-tuning process. 

\vspace{-1mm}
\subsection{Impact of quantization data source}
\label{sec:data_source}
\vspace{-1mm}
The quantized models \textcolor{black}{fine-tuned} with synthetic data achieved very competitive results to the ones \textcolor{black}{fine-tuned} with real data, and even, in many cases, the quantized models \textcolor{black}{fine-tuned} with synthetic data outperformed the quantized model with real data. This is especially true for the 6-bit bandwidth, as shown in Table \ref{tab:result}. For example, using 6-bit bandwidth, the achieved accuracies on LFW by ResNet100 quantized with real and synthetic data are 99.55\% and 99.45\%, respectively.
To illustrate the correlation between the quantized models using real and synthetic data, Figure \ref{fig:act} presents the activation functions value range variables ($[\beta,\alpha]$) of quantized models using real and synthetic data. The high correlation between the activation functions value range of the quantized model using real and synthetic data can be noticed from the overlap in the value range. This indicates that the quantized model is able to capture sufficient data information from the synthetic data to match the FP model output.

\vspace{-1mm}
\section{Conclusion}
\label{sec:conc}
\vspace{-1mm}
This work is the first to explore the potential of regulating the computational cost of existing deep face recognition using low-bit format model quantization in a privacy-friendly process.
In particular, once the model is quantized, synthetically generated face data from unconditional GAN is fed into the FP and quantized model. 
Then, the proposed training paradigm matches the feature embeddings of the FP and quantized model in a normalized embedding space. 
The reported results pointed out the effectiveness of the presented approach in regulating the computational cost of the face recognition model without accessing the original training data or any prior knowledge about the actual data used to train the FP model.

\section*{Acknowledgment}This research work has been funded by the German Federal Ministry of Education and Research and the Hessen State Ministry for Higher Education, Research and the Arts within their joint support of the National Research Center for Applied Cybersecurity ATHENE. This work has been partially funded by the German Federal Ministry of Education and Research (BMBF) through the Software Campus Project.
\bibliographystyle{IEEEtran}
\bibliography{bare_conf}

\newpage






%



\end{document}